\documentclass{article}

\usepackage{microtype}
\usepackage{graphicx}
\usepackage{subfigure}
\usepackage{booktabs} 

\usepackage{hyperref}

\usepackage[accepted]{icml2024}

\usepackage{amsmath}
\usepackage{amssymb}
\usepackage{mathtools}
\usepackage{amsthm}

\usepackage[capitalize,noabbrev]{cleveref}

\usepackage{graphicx}
\usepackage{tabularx}
\usepackage{amsmath,bm}
\usepackage{amssymb}
\usepackage{multirow}
\usepackage{makecell}
\usepackage[toc,page]{appendix}
\newcommand{\RNum}[1]{\uppercase\expandafter{\romannumeral #1\relax}}

\theoremstyle{plain}

\theoremstyle{definition}

\theoremstyle{remark}

\usepackage[textsize=tiny]{todonotes}

\icmltitlerunning{Learning to Reduce: Large Language Models on Structured Data}

\begin{document}

\twocolumn[
\icmltitle{Learning to Reduce: Towards Improving Performance of \\Large Language Models on Structured Data}



\icmlsetsymbol{equal}{*}

\begin{icmlauthorlist}
\icmlauthor{Younghun Lee}{equal,xxx,yyy}
\icmlauthor{Sungchul Kim}{yyy}
\icmlauthor{Ryan Rossi}{yyy}
\icmlauthor{Tong Yu}{yyy}
\icmlauthor{Xiang Chen}{yyy}
\end{icmlauthorlist}

\icmlaffiliation{xxx}{Department of Computer Science, Purdue University, West Lafayette, Indiana, United States}
\icmlaffiliation{yyy}{Adobe Research, San Jose, California, United States}

\icmlcorrespondingauthor{Sungchul Kim}{sukim@adobe.com}

\icmlkeywords{Structured data QA,Table QA,Large Language Models,LLM Prompting,Reinforcement Learning,Policy Optimization}

\vskip 0.3in
]



\printAffiliationsAndNotice{\icmlInternProject}

\begin{abstract}
Large Language Models (LLMs) have been achieving competent performance on a wide range of downstream tasks, yet existing work shows that inference on structured data is challenging for LLMs. 
This is because LLMs need to either understand long structured data or select the most relevant evidence before inference, and both approaches are not trivial.
This paper proposes a framework, Learning to Reduce,
that fine-tunes a language model with On-Policy Learning to generate a reduced version of an input structured data.
When compared to state-of-the-art LLMs like GPT-4, Learning to Reduce not only achieves outstanding performance in reducing the input, but shows generalizability on different datasets. 
We further show that the model fine-tuned with our framework helps LLMs better perform on table QA tasks especially when the context is longer.

\end{abstract}
\section{Introduction}
Recent Large Language Models (LLMs) such as GPT-4 \citep{gpt4}, Llama 2 \citep{touvron2023llama}, and Vicuna \citep{chiang2023vicuna}, have shown the ability to understand language and improve performance on a wide range of downstream tasks \citep{wei2022chain, kojima2022large, wang2022self, yu2022generate, he2024can}. 

Despite their ability, LLMs find it challenging to understand structured data such as knowledge graphs, tables, and databases.
Not only does the structured data have structural dependencies among entities, but it also accompanies long context issues.
The maximum input sequence length of the recent LLMs keeps increasing, yet one cannot put faith in the LLMs' performance when the context is long. 
\citet{liu2023lost} show that the performance of ChatGPT \citep{chatgpt} on a multi-document QA task drops more than $20\%$ when it is prompted with 20 documents and the key information about the answer is located in the middle of the prompt. 
This implies that LLMs are likely to perform unsatisfactory on structured data QA because most structured data tend to be long and the information relevant to the answer can be located in the middle.

To avoid a long context issue of structured data, \citet{jiang2023structgpt} used ChatGPT to identify the most relevant evidence from the structured data before performing QA tasks.
Unfortunately, finding the most relevant evidence was not trivial; experimental results show that around $74\%$ and $28\%$ of the errors came from the incorrect selection of relevant evidence in a KGQA task and a table QA task, respectively.

In this paper, we explore a method to improve LLMs' reasoning ability on structured data by Learning to Reduce; the model efficiently reduces the input structured data by identifying the relevant evidence to the downstream task. 
Specifically, we focus on a table QA as a downstream task, and train a model to generate a table with reduced rows and columns.
Using On-Policy Learning, we fine-tune a T5 language model based on the rewards computed by whether the model selects rows and columns that are relevant to the input question.
Experimental results show that our framework achieves better performance in identifying relevant items from a table QA dataset, WikiTableQuestions \citep{pasupat-liang-2015-compositional}. Additionally, Learning to Reduce outperforms other baseline models including GPT-4 \cite{gpt4} on the generalizability test on an unseen dataset. Lastly, the reduced table generated by our framework helps LLMs perform more accurately on a table QA task, especially when the context is longer.

\begin{figure*}[t!]
    \centering
    \includegraphics[width=0.98\textwidth]{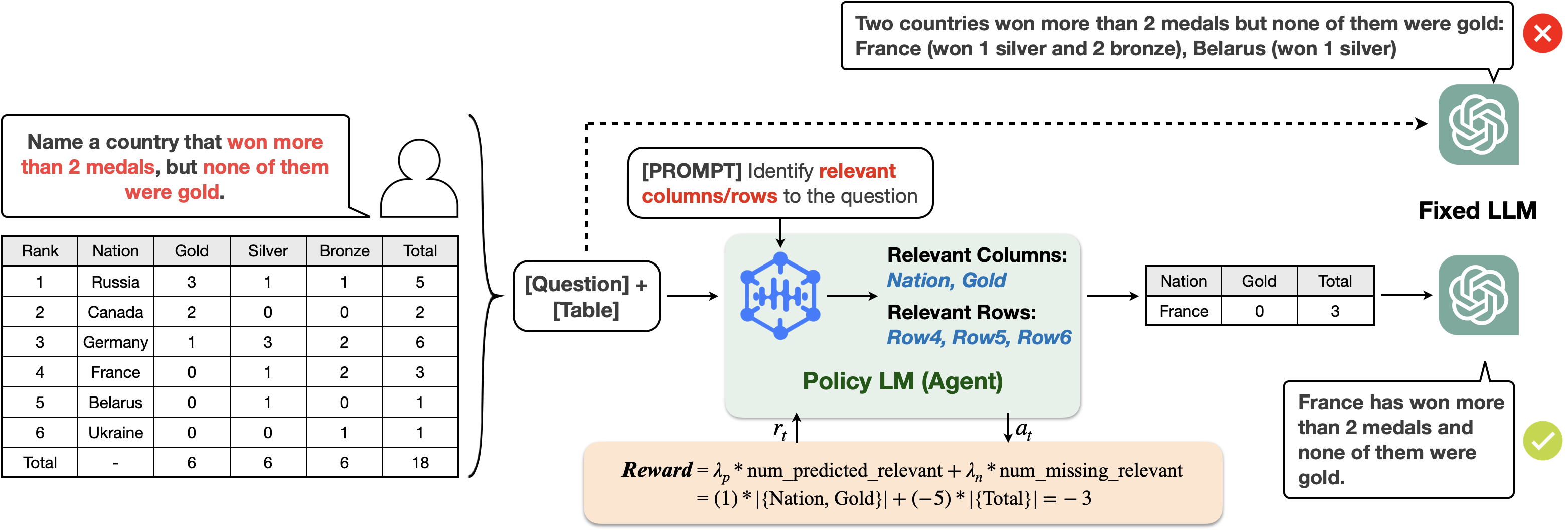}
    \caption{Inference with an original table (dotted arrow) and with a reduced table (solid arrow). Given an input question and a table, a language model (blue hexagon) learns a policy to generate the relevant rows and columns by getting rewards. By learning the optimal policy, our model generates reduced tables which leads the fixed LLM model to perform more accurately on QA tasks.  
    }
    \label{fig:model}
\end{figure*}

To the best of our knowledge, this is the first attempt to train a language model to reduce the input context of structured data. The training framework we suggest is model-agnostic, thus more powerful language models that are trained with our framework can be used as a pre-prompting tool for any structured data QA tasks. This would ultimately maximize the reasoning ability of LLMs as well as the cost efficiency of using them.

\section{Learning to Reduce}
We design a training framework for a language model that learns to generate the relevant evidence for a given structured data QA instance.

Formally, we define an input space that consists of an input context $\bm{c}$, a task description $\bm{x}$, and an output $\bm{y}$. In the table QA task, for instance, $\bm{c}$ is an input table, $\bm{x}$ is an input question, and $\bm{y}$ is an answer to the question. We use a heuristics $h(\cdot)$ that identifies a subset of the context that is relevant to the task, namely $\bm{c_r}=h(x,c)$. Our framework trains a language model, $\theta(\bm{z|x,c})$, that generates a reduced input context $\bm{z}$, using a target reduced input context $\bm{c_r}$. Our final objective is to help LLMs better perform on the task, thus we prompt an LLM with reduced context $\bm{z}$ to generate an output. Denoting $\psi_{LLM}$ as a fixed LLM model used for a table QA task, the predicted output $\bm{\hat{y}}$ follows $\bm{\hat{y}} \sim \psi_{LLM}(\cdot | \bm{x, z})$.
Figure \ref{fig:model} illustrates how our proposed model works.






\subsection{Language Model as a Policy Network}
\label{sec:language-model-as-policy-network}
We consider a language model as a policy network. Following an approach proposed by \citet{li2023guiding}, we first fine-tune the language model parameters before applying policy learning objectives. 

Supervised fine-tuning requires a dataset that has relevant rows and columns annotated for input questions and tables. To automatically identify relevant items, we use a table QA dataset which has a text-to-SQL annotation. The heuristics for automatically identifying relevant rows and columns is to execute SQL queries while iteratively removing rows and columns one by one from the input tables; if executing a SQL query can generate an answer on a table even after removing some rows and columns, then the removed items are irrelevant to the input question.

Using the WikiTableQuestions (WTQ) dataset \citep{pasupat-liang-2015-compositional} and its corresponding text-to-SQL annotation from SQUALL \citep{shi-etal-2020-potential}, we identify relevant rows and columns using the aforementioned heuristics. WTQ dataset was selected because the questions in the dataset are simple enough to be translated into SQL queries, yet the state-of-the-art LLMs' inference ability on this dataset is not reliable compared to other widely used table QA datasets\footnote{\citet{jiang2023structgpt} reported ChatGPT's accuracy on WTQ is 43.3, while it achieves 51.6 on WikiSQL \citep{zhong2017seq2sql} and 82.9 on TabFact \citep{chen2019tabfact}}.
Appendix \ref{appendix:data} describes the details of the data.

We use a sequence-to-sequence language model with FLAN-T5-Large \citep{chung2022scaling} pre-trained checkpoint as a base model. For better optimization, we train two models separately, one for the column reduction and the other for the row reduction. The two models work in sequence; the column reduction language model is applied and generates a table with relevant columns, then the row reduction model gets a column-reduced table and determines relevant rows. Thus the column reduction model gets all column headers in the prompt, whereas the row reduction model gets rows only with relevant columns. More details of the model choice and prompt designs are described in Appendix \ref{appendix:experiments}. 



\subsection{Policy Optimization}
\label{sec:policy-optimization}
We consider fine-tuned language models as initial policy network parameters and further optimize the models by maximizing the rewards. The models get positive rewards for selecting the correct relevant items, and get two types of negative rewards. 

Type \RNum{1} errors occur when the models generate rows and columns that are not relevant. Having this error is not preferred but it is not critical; 
when the policy network generates a table with some irrelevant rows and columns, LLMs can still perform a QA task as long as the relevant items are in the table. In this case, the policy network gets a small negative reward.

On the other hand, type \RNum{2} errors need to be considered more seriously, as this happens when the models fail to generate relevant rows and columns. In this case, LLMs would have to perform inference without the necessary information and fail to correctly answer the question. We consider this as a critical error and the policy network gets a high negative reward. 

To formally define a reward function, consider $\bm{z}$ and $\bm{c_r}$ as a concatenated text of rows and columns from a set $Z$ and $C_r$. For instance, $Z=\{ \text{col1}, \text{row1}\}$ when $\bm{z}$ is ``col1, row1''. The model maximizes the reward $\mathcal{R}$, which is defined as:

\vskip -0.1in
$$ \mathcal{R} = \lambda_p|Z\cap C_r|+\lambda_{n1}|Z-C_r|+\lambda_{n2}|C_r-Z|$$

$\lambda_p>0$, $\lambda_{n1}<0$, $\lambda_{n2}\ll0$ are coefficients for correct predictions, type \RNum{1} errors, and type \RNum{2} errors, respectively. Details of the policy network architecture, implementation, and reward optimization are described in Appendix \ref{appendix:model}.




\section{Experiments}

\textbf{Baselines:} First, we fine-tune a RoBERTa token classifier \citep{liu2019roberta} as a baseline. 
Similar to supervised fine-tuning of a language model, input is given with question and column/row values, and the token classifier models generate tags for each column/row whether they are relevant or not. 
Another baseline is a zero-shot GPT-4 model\footnote{We use version gpt-4-0613}. In this setting, we provide input questions with column/row values, then ask the GPT-4 model to select relevant rows and columns. Lastly, we use fine-tuned FLAN-T5 models as a baseline. This is to measure the effectiveness of the On-Policy Learning component in context reduction.

\begin{table}[t]\footnotesize
\caption{\label{tab:context-reduce-experiments}
Input context reduction tested on WTQ test set (WTQ) and on an unseen test set (HybQA).
}
\vskip 0.1in
\centering
\begin{tabularx}{0.95\columnwidth}{X|r|r}
\hline
\multicolumn{3}{c}{\textbf{Recall on Context Reduction}}\\
\hline
\hline
\textbf{Column Reduction} & \textbf{WTQ} & \textbf{HybQA} \\
\hline
RoBERTa Token Clf & 89.08 & 77.32\\
Zero-shot GPT-4 & 74.03 & 71.48\\
Fine-tuned FLAN-T5 & 90.19 & 82.72\\
$\pmb{\star}$ Learning to Reduce (\textit{ours}) & \textbf{\textit{91.82}}& \textbf{\textit{87.22}}\\
\hline
\hline
\textbf{Row Reduction} & \textbf{WTQ} & \textbf{HybQA} \\
\hline
RoBERTa Token Clf & 95.06 & 60.35 \\
Zero-shot GPT-4 & 86.13 & 92.12\\
Fine-tuned FLAN-T5 & 95.21 & 90.22\\
$\pmb{\star}$ Learning to Reduce (\textit{ours}) & \textbf{\textit{96.17}} & \textbf{\textit{93.78}}\\
\hline
\end{tabularx}

\vskip -0.24in
\end{table}

\textbf{Evaluation Metrics:} As described in \ref{sec:policy-optimization}, context reduction models are considered better when they generate relevant items as many as possible, even though their generation includes irrelevant items. Thus we use \textbf{recall} scores as a metric for context reduction and measure how many relevant rows and columns are selected from the models. More discussions and results of the precision scores are reported in Appendix \ref{appendix:downstream}.

\textbf{Dataset:} The models are tested on the WTQ test set as well as on Hybrid QA \citep{chen-etal-2020-hybridqa}. Hybrid QA is a table QA task that not only requires reasoning on a given table but also needs more complex language understanding. We test the models on Hybrid QA to evaluate their generalizability. The model needs to be robust to different data distributions because fine-tuning is not always possible. Thus we argue that the model's performance on Hybrid QA is more significant when we consider deploying the model for more general usages. Table \ref{tab:context-reduce-experiments} shows the experimental results. 

\subsection{Performance and Generalizability}
RoBERTa token classifier outperforms zero-shot GPT-4 on both column and row reduction. However, the model's performance significantly drops when it is tested on Hybrid QA; the column reduction model shows a $12\%$ drop in column reduction and a $35\%$ drop in row reduction. This result shows that fine-tuned BERT-based token classifiers lack generalizability.


GPT-4 is a very large and highly contextual model, yet it has difficulty understanding the relatedness between a table and a question. This result aligns with the selection error statistics in the existing research \citep{jiang2023structgpt}. As the model is not fine-tuned on a specific dataset, its performance on WTQ and Hybrid QA are similar as expected. A slight drop from $74\%$ to $71.5\%$ on column reduction implies that Hybrid QA is a little more challenging dataset than WTQ.

Regarding the language model approaches, On-Policy Learning helps improve performance. Not only does the Learning to Reduce outperform other baselines on the WTQ test set, but it also exhibits remarkable results on Hybrid QA.
This result implies that policy networks can capture more general knowledge about context reduction and opens the possibility of applying Learning to Reduce on other structured data QA tasks without fine-tuning again. 

\begin{figure}[t!]
    \centering
    \includegraphics[width=0.98\columnwidth]{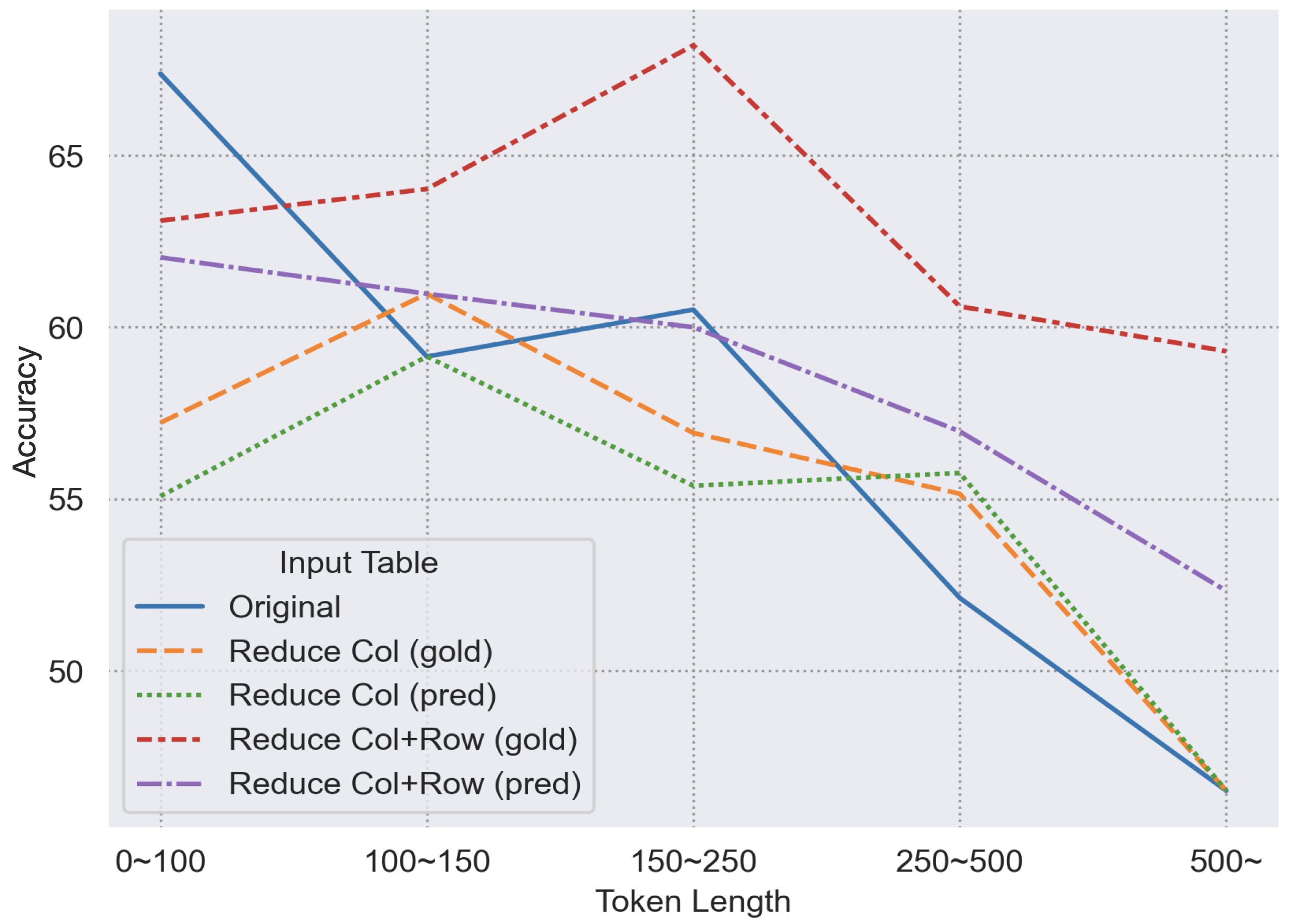}
    \caption{Accuracy (precision) of GPT-4 model on WTQ test set with different input context tables. Reducing both rows and columns (red and purple) is more powerful when the context is longer.
    }
    \label{fig:downstream-task}
    \vskip -0.1in
\end{figure}

\subsection{Downstream Task Performance}
The ultimate goal of our framework is to help LLMs better perform table QA tasks with reduced rows and columns.
Figure \ref{fig:downstream-task} illustrates the accuracy of the GPT-4 model in answering the WTQ questions when prompted with different context inputs\footnote{We compute precision (i.e. how many GPT-4 predictions are correct out of all prompted instances)}. Note that the x-axis is the number of tokens of the input tables before context reduction.

Our hypothesis, that LLMs lack inference ability on structured data due to its long context, is validated by the results of original input table; the accuracy of the original input drops as the table gets longer (blue line in Figure \ref{fig:downstream-task}). 

We assumed that the input table representations with only columns reduced can moderately improve the inference ability of LLMs. This is because the majority of tables in WTQ datasets are long-narrow tables (i.e. fewer columns than rows), thus removing a few columns would reduce a great number of input tokens. However, the green and orange line shows that reducing only columns does not improve the overall performance of LLMs. 

The performance of both rows and columns reduced (red and purple line), on the other hand, is significantly more stable, especially when the original input table is longer. 
The purple line shows that our proposed method provides better QA results compared to pure LLM-based approaches (blue), even though the policy network's recall on context reduction is not 100\% accurate. Our proposed method proves to be more impactful for maximizing LLMs' ability to understand long structured data. More results with a different LLM, GPT-3.5-turbo, are described in Appendix \ref{appendix:downstream}.

\section{Related Work}
\textbf{LLM on Structured Data}: Various approaches have been proposed to integrate structured data into LLM prompts. \citet{ye2023large} proposed DATER which uses LLMs to decompose inputs into sub-questions/-tables, then executes queries to improve the reasoning ability of the LLM. \citet{hegselmann2023tabllm} converted tables into natural language and fine-tuned LLMs on downstream tasks. StructGPT \citep{jiang2023structgpt} iteratively gets the most relevant evidence from structured data, then prompts LLMs for better inference.

\textbf{LLM Prompt Engineering}: Another direction to improve the downstream task performance of a fixed LLM is to optimize the prompts. GrIPS \citep{prasad-etal-2023-grips} uses a gradient-free, edit-based instruction search method. 
AutoPrompt \citep{shin-etal-2020-autoprompt} and RLPrompt \citep{deng-etal-2022-rlprompt} train the model to generate tokens that help LLMs better perform on downstream tasks. Directional Stimulus Prompting \citep{li2023guiding} trains the model to generate keywords or hints that can directly help LLMs to accomplish downstream tasks.

\section{Discussion and Conclusion}
There are a few areas that can further improve the paper. The paper would benefit from experiments on more table QA datasets; the baseline models as well as Learning to Reduce are only tested on Hybrid QA datasets for measuring the models' generalizability. Similarly, the downstream task performance results would be more strongly supported by experimenting with more table QA datasets---measuring whether LLMs' inference ability improve or not on other table QA datasets after context reduction.



The goal of this paper can be stretched to develop a framework that improves performance on LLMs on any type of structured data, including knowledge base and databases. Testing the framework with other types of structured data remains as future work.

Lastly, our framework can benefit from adding task-specific rewards. There are some cases where our model generates a very small number of rows and columns. Even though they are correct reduction, LLMs end up being confused with the input with too much reduced rows and columns. If there are more reward signals from how LLMs perform on downstream tasks with the reduced context input, the model would be able to learn better representations. We leave this as future work.

To conclude, we propose Learning to Reduce, a framework that learns to generate a reduced context. The framework adopts a novel reward function for context reduction and fine-tunes a language model through On-Policy Learning. Not only does our framework outperform a state-of-the-art LLM (e.g. GPT-4) in context reduction, but it is also generalizable to an unseen dataset. We further show that the output of our framework improves the LLM's performance on downstream tasks. Learning to Reduce is model-agnostic thus it can be applied to different types of structured data such as knowledge base and databases in future studies.
\bibliography{example_paper}
\bibliographystyle{icml2024}

\newpage
\appendix

\section{Data Statistics}
\label{appendix:data}

The original WTQ dataset consists of around 14K, 3.5K, 4.5K instances for training, validation, and test set. When we use the SQUALL annotation and identify the instances with valid SQL query annotation, the number of remaining instances is 10K, 3K, 1.5K for each of the data split. Detailed statistics of the dataset is described in Table \ref{tab:data-statistics}.

From the observation, most of the tables in the dataset are long-narrow tables where there are more rows than columns. The average number of tokens of input context table is 1,333, and there are around 300 isntances that exceed the maximum sequence length of recent GPT checkpoints, 4,096.

\section{Experiment Details}
\label{appendix:experiments}
All models are trained and tested on 8-core NVIDIA Tesla A10 GPU with 24GB RAM. Policy networks are trained over 10 iterations and evaluated for every 3 iterations. The total amount of policy network training took 18 hours. 

We design our framework to fine-tune two separate language models, one for the column reduction and the other for the row reduction. One of the reasons we did not train a single model for context reduction is to minimize repetition. Consider an input table, and denote the number of all columns $N$, number of relevant columns $n$, number of all rows $M$, and number of relevant rows $m$. When we have a language model with the maximum input sequence length as $L$, the single model needs to be prompted $\lceil\frac{N\times M}{L}\rceil$ times. When we reduce the columns first, the number of prompting reduces to $\lceil\frac{n\times M}{L}\rceil+1$ times. $+1$ represents language model prompting for column reduction, and this can be done within a single prompting because the maximum number of columns is 25 in WTQ dataset. Another reason is to make the policy network optimize better. The single language model that generates relevant rows and columns at the same time fails to be fine-tuned and only generates columns or rows even after training the model with several iterations.

For context reduction, language models are prompted as \textit{``Select relevant columns from a table to answer a question. Output `@' if done generating. Question: \{Table QA question\}, List of column headers: \{Col1, Col2, ..\}''}. For row reduction, the rows are represented with each cell's corresponding column value as \textit{``Select relevant rows from a table to answer a question. Output `@' if done generating. Question: \{Table QA question\}, List of rows in a format `rowX: (column name=value)': \{row1: (col1=val1), row2: (col2=val2), ..\}''}. When there are a lot of rows in the table and representing them exceeds the maximum token limit of the language model, we truncate the table and prompt the model. Thus in some cases, the language model can only generate the end-of-sequence token when the given sub-table does not contain the relevant evidence.

For the generalizability test, we manually annotate 200 instances of the Hybrid QA test set in order to identify the relevant rows and columns for a given question and table pair, because Hyrbid QA does not have corresponding SQL queries annotated for the instances.

\begin{table}\footnotesize
\caption{\label{tab:data-statistics}
Statistics of the WTQ dataset with valid SQUALL annotation
}
\vskip 0.1in
\centering
\begin{tabularx}{0.95\columnwidth}{X|r}
\hline
\multicolumn{2}{c}{\textbf{WTQ with valid SQUALL annotation}} \\
\hline
\# of train / valid / test set & 10K / 3K / 1.5K \\
\# of unique tables & 5.4K \\
\# of questions per table & 2.73 \\
\hline
Avg / Max \# of columns & 9 / 25\\
Avg / Max \# of rows & 46 / 753\\
Avg / Max \# of cells & 230 / 3,832\\
\hline
Avg / Max \# of context tokens & 1,333 / 20,324\\
\# instances $>$ 4,096 tokens & 248\\
\# instances $>$ 8,192 tokens & 60\\
\hline
\end{tabularx}

\end{table}

\section{Model Architecture}
\label{appendix:model}
The initial policy network parameters are pre-trained language model that is small enough to fine-tune. We use a sequence-to-sequence language model with FLAN-T5-Large \citep{chung2022scaling} checkpoint as a base model and fine-tune the model with the annotation of relevant rows and columns for each question and table pair.
Mathematically, we fine-tune the language models by maximizing the log-likelihood as follows:

$$\mathcal{L}_{FT} = -\mathbb{E}\text{log}\theta_{LM}(\bm{z|x,c})$$

We further tune the language model by computing the rewards from the model's predictions on relevant rows and columns. The model gets positive rewards for selecting the correct relevant items, and gets negative rewards for incorrect selection. The model is penalized more when it does not select the relevant items compared to when it selects irrelevant items. As described in \ref{sec:policy-optimization}, the reward is defined as:

$$ \mathcal{R} = \lambda_p|Z\cap C_r|+\lambda_{n1}|Z-C_r|+\lambda_{n2}|C_r-Z|$$

Note that $C_r$ is derived from a heuristics with text-to-SQL annotation, namely $\bm{c_r} = h(x,c)$. To formally define the parameter update condition, we aim to maximize the following objective:
$$\text{max}_{\theta_{LM}}\mathbb{E}_{\bm{z}\sim \theta_{LM}(\cdot|\bm{x,c})}[\mathcal{R}(\bm{x,c, z})]$$

To make the optimization tractable for policy network, we employ proximal policy optimization (PPO) method \citep{schulman2017proximal}. We consider a fine-tuned language model as an initial policy network (i.e., $\pi_0=\theta_{LM}$) and update the policy network $\pi$ using PPO. The policy network's generation of relevant information can be considered as a Markov Decision Process $\langle \mathcal{S}, \mathcal{A}, r, \mathcal{P} \rangle$, where $\mathcal{S}$ is a state space, $\mathcal{A}$ is an action space, $r$ is a reward function, and $\mathcal{P}$ is transition probabilities. At each time step $t$ in an episode, the model selects an action (i.e., generating tokens of relevant information) from $\mathcal{A}$, based on the state of the current time step. The state at time $t$ is defined with the input and the policy network's previous generations, that is, $\bm{z_{t}} = \pi(\cdot|\bm{x, c, z_{<t-1}})$. The episode ends when the policy network generates an end-of-sequence token.

Following the existing RL approaches on NLP applications \citep{ziegler2019fine}, we employ KL divergence penalty rewards that dynamically adapt the coefficient $\beta$ in different time steps to minimize excessive parameter updates from the initial policy. The action space for token generation often stretches to the size of vocabulary and it makes optimization costly. To minimize such issues, \citep{ramamurthy2022reinforcement} proposed NLPO, an approach that masks out the least probable tokens using top-$p$ sampling. We set $p$ as 0.9 in the experiments. The policy network's reward function $r\bm{(x, c, z)}$ is defined as:

$$r(\bm{x,c, z}) = \mathcal{R}(\bm{x,c, z}) - \beta\log{\frac{\pi(\bm{z|x,c})}{\theta(\bm{z|x,c})}}$$
$$\bm{e}_t = \text{CLIP}\left(\frac{\text{KL}(\pi_t, \theta)-\text{KL}_{\text{target}}}{\text{KL}_{\text{target}}} , -0.2, 0.2 \right)$$
$$\beta_{t+1} = \beta_t(1+K_{\beta}\bm{e}_t)$$

\section{Additional Results}
\label{appendix:downstream}

\begin{table}[t]\footnotesize

\caption{\label{tab:context-reduce-experiments-precision}
Precision scores of input context reduction tested on WTQ test set (WTQ) and on an unseen test set (HybQA).
}
\vskip 0.1in

\centering
\begin{tabularx}{0.95\columnwidth}{X|r|r}
\hline
\multicolumn{3}{c}{\textbf{Precision on Context Reduction}}\\
\hline
\hline
\textbf{Column Reduction} & \textbf{WTQ} & \textbf{HybQA} \\
\hline
RoBERTa Token Clf & 93.32 & 77.61\\
Zero-shot GPT-4 & 72.13 & 86.78\\
Fine-tuned FLAN-T5 &93.30  & 75.11\\
$\pmb{\star}$ Learning to Reduce (\textit{ours}) & 90.32 & 75.15\\
\hline
\hline
\textbf{Row Reduction} & \textbf{WTQ} & \textbf{HybQA} \\
\hline
RoBERTa Token Clf & 92.95& 7.93\\
Zero-shot GPT-4 &  87.91& 94.35\\
Fine-tuned FLAN-T5 &  89.01& 98.22\\
$\pmb{\star}$ Learning to Reduce (\textit{ours}) & 91.12& 97.88\\
\hline
\end{tabularx}

\end{table}

\subsection{Context Reduction Precision}
Table \ref{tab:context-reduce-experiments-precision} shows the precision scores of models on context reduction. We use recall scores for evaluating different models, yet precision scores cannot be ignored. For instance, if a model chooses all columns and rows as relevant for all cases, it achieves 100\% recall but the model would not be considered as well trained.

Learning to Reduce does not outperform other models with regard to the precision scores. When it is compared to its non-RL counterpart, fine-tuned FLAN-T5, the precision scores are almost similar or decreasing. This is expected because our framework is tuned with recall-based reward. As our framework does not show exceptionally poor performance on precision scores compared to other baselines, we argue that Learning to Reduce does not achieve its outperforming recall scores by simply selecting more columns and rows.

\subsection{Downstream Task Performance}

\begin{figure}[t!]
    \centering
    \includegraphics[width=0.98\columnwidth]{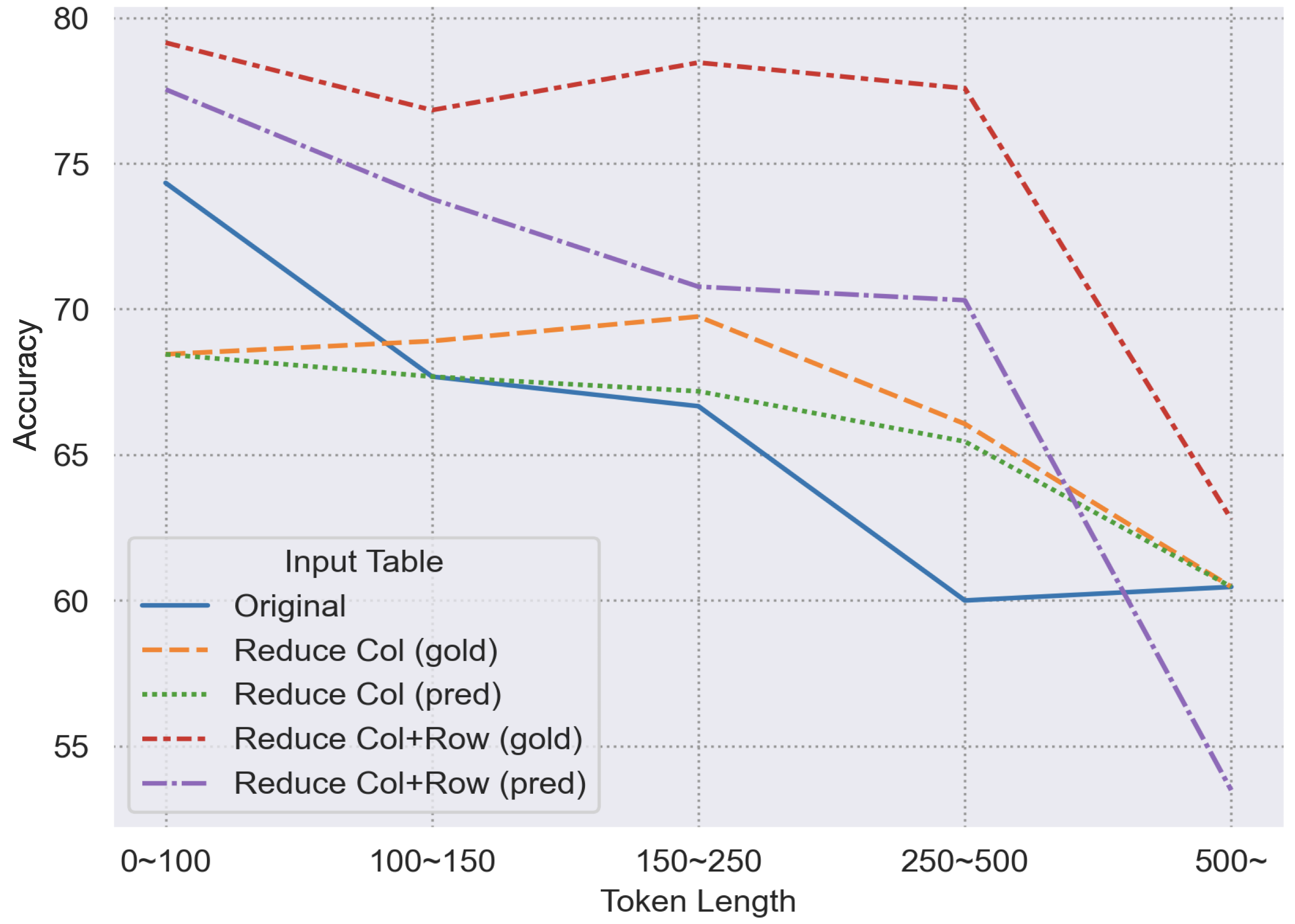}
    \caption{Accuracy (precision) of GPT-3.5-turbo on WTQ test set with different input context tables. 
    }
    \label{fig:downstream-task-addtl}
\end{figure}
Figure \ref{fig:downstream-task-addtl} shows table QA performance on WTQ dataset when we use a different LLM, GPT-3.5-turbo. Similar to the results using GPT-4 model, LLM's inference ability drops as the input context gets longer. We can also see the same trend for inputs with reduced context---input tables with only columns reduced does not improve the LLM's performance while input tables with both rows and columns reduced leads LLMs perform better. The gap between the gold and predicted context reduction motivates the necessity of improving the accuracy of context reduction in future studies.

\end{document}